\documentclass{article}

\PassOptionsToPackage{numbers, compress}{natbib}

\usepackage[preprint]{neurips_2020}

\usepackage[utf8]{inputenc} 
\usepackage[T1]{fontenc}    
\usepackage{hyperref}       
\usepackage{url}            
\usepackage{booktabs}       
\usepackage{amsfonts}       
\usepackage{nicefrac}       
\usepackage{microtype}      
\usepackage{amsmath}
\usepackage{graphicx}
\usepackage{cleveref}
\usepackage{algorithm}
\usepackage{algpseudocode}
\usepackage{wrapfig}
\usepackage{tabularx}   
\usepackage{enumitem}  

\def\x{\mathcal{I}}
\def\y{\mathbf{y}}

\def\EPE{\text{EPE}}
\def\fovea{f_\text{fovea}}  
\def\rimg{{r_\text{img}}}
\def\gavp{{g_\text{AVP}}}
\def\ndata{{N^{\text{data}}}}

\newcommand{\PSNOGS}{PS-NOGS}
\newcommand{\RAM}{RAM}
\newcommand{\RAMP}{RAM+}

\newcommand{\HGS}{PS-HGS}
\newcommand{\PSHO}{PS-H1}
\newcommand{\PSHF}{PS-H5}

\newcommand{\ourparagraph}[1]{\textbf{#1}~~}

\newcommand{\lref}[1]{\cref{#1}}   
\AtBeginDocument{%
  \crefname{section}{Section}{Sections}%
  \crefname{figure}{Fig.}{Figures}%
  \crefname{appendix}{Appendix}{Appendices}%
}

\newcommand{\beq}[1]{\begin{equation} \eqlab{#1}}
\newcommand{\eeq}{\end{equation}}
\newcommand{\bal}[1]{\begin{align}#1}
\newcommand{\eal}[1]{\end{align}}

\newcommand{\bsub}{\begin{subequations}}
\newcommand{\esub}{\end{subequations}}

\newcommand{\norm}[1]{\left\lVert#1\right\rVert}

\def\argmin{\mathop{\rm arg\,min}}

\newcommand{\EX}{\mathbb{E}}

\newcommand{\entropy}{\mathcal{H}}

\newenvironment{rcases}{\left.\begin{aligned}}{\end{aligned}\right\rbrace}

\usepackage{todonotes}  
\usepackage{xcolor}  

\title{Near-Optimal Glimpse Sequences for Improved \\ Hard Attention Neural
  Network Training}

\author{%
  William Harvey \\
  Department of Computer Science\\
  University of British Columbia\\
  \texttt{wsgh@cs.ubc.ca} \\
  \And
  Michael Teng \\
  Department of Engineering\\
  University of Oxford\\
  \texttt{mteng@robots.ox.ac.uk} \\
  \And
  Frank Wood \\
  Department of Computer Science\\
  University of British Columbia\\
  \texttt{fwood@cs.ubc.ca} \\
}

\begin{document}
\maketitle

\begin{abstract}
  Hard visual attention is a promising approach to reduce the computational
  burden of modern computer vision methodologies. Hard attention mechanisms are
  typically non-differentiable. They can be trained with reinforcement learning
  but the high-variance training this entails hinders more widespread
  application. We show how hard attention for image classification can be framed
  as a Bayesian optimal experimental design (BOED) problem. From this
  perspective, the optimal locations to attend to are those which provide the
  greatest expected reduction in the entropy of the classification distribution.
  We introduce methodology from the BOED literature to approximate this optimal
  behaviour, and use it to generate `near-optimal' sequences of attention
  locations. We then show how to use such sequences to partially supervise, and
  therefore speed up, the training of a hard attention mechanism. Although
  generating these sequences is computationally expensive, they can be reused by
  any other networks later trained on the same task.
\end{abstract}




\section{Introduction}
\label{sec:introduction}
Attention can be defined as the ``allocation of limited cognitive processing
resources''~\cite{anderson2005cognitive}. In humans the density of
photoreceptors varies across the retina. It is much greater in the
centre~\cite{bear2007neuroscience} and covers an approximately 210 degree field
of view~\cite{traquair1949introduction}. This means that the visual system is a
limited resource with respect to observing the environment and that it must be
allocated, or controlled, by some attention mechanism. We refer to this kind of
controlled allocation of limited sensor resources as ``hard'' attention. This is
in contrast with ``soft'' attention, the controlled application of limited
computational resources to full sensory input. Our focus in this paper is on
hard attention mechanisms. Their foremost advantage is the ability to solve
certain tasks using orders of magnitude less sensor bandwidth and computation
than the alternatives~\cite{katharopoulos2019processing, rensink2000dynamic}.

\begin{figure}[t]
  \centering
  \includegraphics[scale=1]{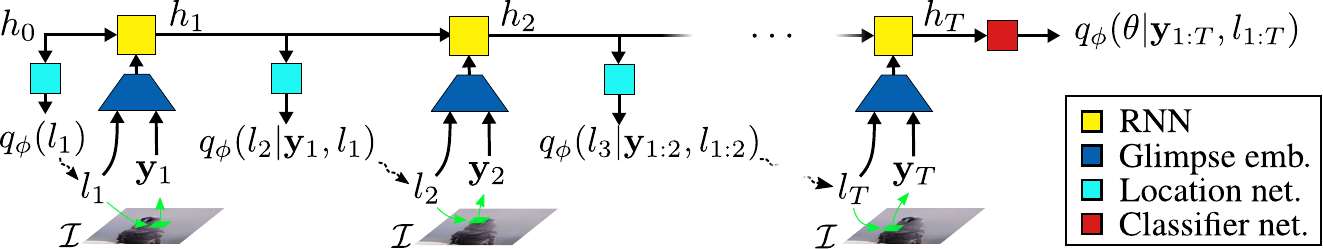}
  \caption{The hard attention network architecture we consider, consisting of an
    RNN core (yellow), a location network (light blue), a glimpse embedder (dark
    blue), and a classifier (red). $h_t$ is the RNN hidden state after $t$
    steps. %
    The network outputs distributions over where to attend ($l_t$) at each time step, and over the class label ($\theta$) after $T$ steps.
    %
  }
  \label{fig:hard-attention-architecture}
\end{figure}

This paper focuses on the application of hard attention in image classification.
%
%
Our model of attention (shown in \lref{fig:hard-attention-architecture}) is as
follows: a recurrent neural network (RNN) is given $T$ steps to classify some
unchanging input image. Before each step, the RNN outputs the coordinates of a
pixel in the image. A patch of the image centered around this pixel is then fed
into the RNN. We call this image patch a glimpse, and the coordinates a glimpse
location.
As such, the RNN controls its input by selecting each glimpse location, and this
decision can be based on previous glimpses.
After $T$ steps, the RNN's hidden state is mapped to a classification output. As
with most artificial hard attention
mechanisms~\cite{mnih2014recurrent,ba2014multiple}, this output is not
differentiable with respect to the sequence of glimpse locations selected. This
makes training with standard gradient backpropagation impossible, and so high
variance gradient estimators such as REINFORCE~\cite{williams1992simple} are
commonly used instead~\cite{mnih2014recurrent,ba2014multiple}. The resulting
noisy gradient estimates make training difficult, especially for large $T$.


In order to improve hard attention training, we take inspiration from
neuroscience literature which suggests that visual attention is directed so as
to maximally reduce entropy in an agent's world model~\cite{bruce2009saliency,
  itti2009bayesian, schwartenbeck2013exploration, feldman2010attention}. There
is a corresponding mathematical formulation of such an objective, namely
Bayesian optimal experimental design (BOED) \cite{chaloner1995bayesian,
  foster2019variational}. BOED tackles the problem of designing an experiment to
maximally reduce uncertainty in some unknown variable. In the case of hard
visual attention, the `experiment' is the process of taking a glimpse; the
`design' is the glimpse location; and the unknown variable is the class label.
In general, BOED is applicable only when a probabilistic model of the experiment
exists. This could be, for example, a prior distribution over the class label and a
generative model for the observed image patch conditioned on the class label and
glimpse location. We leverage generative adversarial networks
(GANs)~\cite{goodfellow2014generative,karras2018style,singh2019finegan} to
provide such a model.

We use methodology from BOED to introduce the following training procedure for
hard attention networks, which we call partial supervision by near-optimal
glimpse sequences (\PSNOGS{}).
\begin{enumerate}[itemsep=0pt, topsep=0pt]
\item We assume that we are given an image classification task and a
  corresponding labelled dataset. Then, for some subset of the training images,
  we determine an approximately optimal (in the BOED sense) glimpse location for
  a hard attention network to attend to at each time step. We refer to the
  resulting sequences of glimpse locations as near-optimal glimpse sequences.
  \Cref{sec:nogs} describes our novel method to generate them.
\item We use these near-optimal glimpse sequences to give an additional
  supervision signal for training a hard attention network.
  We introduce our novel training objective for this in
  \Cref{sec:partially-supervised-training}, which utilises training images both
  with and without such sequences.
\end{enumerate}
We empirically investigate the performance of \PSNOGS{} and find that it speeds
up training compared to our baselines, and leads to qualitatively different
behaviour with competitive accuracy. We also validate the use of BOED to generate
the glimpse sequences by showing that even partial supervision by hand-crafted
glimpse sequences does not have such a beneficial effect on training.

\section{Hard attention} \label{sec:hard-attention}
Given an image, $\x$, we consider the task of inferring its label, $\theta$. We
use an architecture based on that of \citet{mnih2014recurrent}, shown in
\lref{fig:hard-attention-architecture}. It runs for a fixed number of steps,
$T$. At each step $t$,
the RNN samples a glimpse location, $l_t$, which is conditioned on the previous
glimpses via the RNN's hidden state. A glimpse, in the form of a contiguous
square of pixels, is extracted from the image at this location. We denote this
$\y_t = \fovea(\x, l_t)$. An embedding of $\y_t$ and $l_t$ is then input to the
RNN.
After $T$ glimpses, the network outputs a classification distribution
$q_\phi(\theta|\y_{1:T},l_{1:T})$, where $\phi$ are the learnable network
parameters. \citet{mnih2014recurrent} use glimpses consisting of three image
patches at different resolutions, but the architectures are otherwise identical.

During optimisation, gradients cannot be computed by simple backpropagation
since $\fovea$ is non-differentiable. An alternative, taken by
\citet{mnih2014recurrent} and others in the
literature~\cite{ba2014multiple,sermanet2014attention}, is to obtain
high-variance gradient estimates using REINFORCE~\cite{williams1992simple}.
Although these are unbiased, their high-variance has made scaling beyond simple
problems such as digit classification~\cite{netzer2011reading} challenging. Much
research has focused on altering the architecture to ease the learning task: for
example, many process the full, or downsampled, image before selecting glimpse
locations~\cite{ba2014multiple,sermanet2014attention,katharopoulos2019processing,elsayed2019saccader}.
We summarise these innovations in \lref{sec:related-work} but note that they
tend to be less suitable for low-power computation.
%
%
We therefore believe that improved training of the architecture in
\cref{fig:hard-attention-architecture} is an important research problem, and it
is the focus of this paper.

\section{Bayesian optimal experimental design} \label{sec:boed} Designing an
experiment to be maximally informative is a fundamental problem that applies as
much to tuning the parameters of a political survey~\cite{warwick1975sample} as
to deciding where to direct attention to answer a query.
BOED~\cite{chaloner1995bayesian} provides a unifying framework for this by
allowing a formal comparison of possible experiments under problem-specific
prior knowledge.
Consider selecting the design, $l$, of an experiment to infer some unknown
parameter, $\theta$. For example, $\theta$ may be the median lethal dose of a
drug, and $l$ the doses of this drug given to various groups of
rats~\cite{chaloner1995bayesian}. Alternatively, as we consider in this paper,
$\theta$ is the class label of an image and $l$ determines which part of the
image we observe.
The experiment results in a measurement of $\y \sim p(\y|l, \theta)$. For
example, $\y$ could be the number of rats which die in each group or the
observed pixel values. Given a prior distribution over $\theta$ and knowledge of
$p(\y|l, \theta)$, we can use the measurement to infer a posterior distribution
over $\theta$ using Bayes' rule: $p(\theta|\y,l) = \frac{p(\y|l,
  \theta)p(\theta)}{\int p(\y|l, \theta)p(\theta) \mathrm{d}\theta}$.

\begin{figure}
  \centering
  \includegraphics[scale=1]{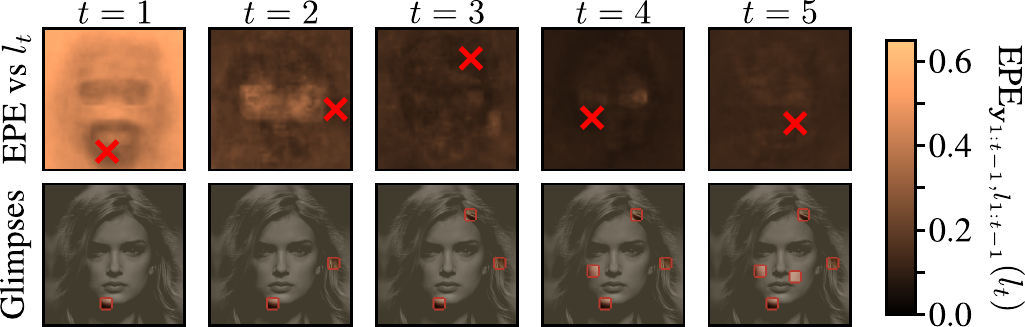}
  \caption{A near-optimal glimpse sequence being generated for the task of
    inferring the attribute `Male'. \textbf{Top row:} A heatmap of estimated
    expected posterior entropy for each possible next glimpse location $l_t$.
    The red cross marks the minimum, which is chosen as the next glimpse
    location. \textbf{Bottom row:} Observed parts of the image after taking each
    glimpse. }
  \label{fig:epe-maps}
\end{figure}

The aim of our experiment is to infer $\theta$, and so a well designed
experiment will reduce the uncertainty about $\theta$ by as much as possible.
The uncertainty after the experiment can be quantified by the Shannon entropy in
the posterior:
\begin{equation}
  \label{eq:pe}
  \entropy \left[ p(\theta | \y, l) \right] = \EX_{p(\theta | \y, l)} \left[ - \log p(\theta | \y, l) \right].
\end{equation}
To maximally reduce the uncertainty, we wish to select $l$ to minimise this
posterior entropy. However, the design of the experiment must be chosen before
$\y$ is measured and so we cannot evaluate the posterior entropy exactly.
Instead, we minimise an expectation of it over $p(\y|l)=\EX_{p(\theta)}\left[
  p(\y|l, \theta) \right]$, the marginal distribution of $\y$. This is the
expected posterior entropy, or $\EPE$:
\begin{align}
  \label{eq:epe}
  \text{EPE} (l) &= \EX_{p(\y|l)} \left[ \entropy \left[ p(\theta | \y, l) \right]  \right].
\end{align}

Above, we considered the case of selecting a one-off design for an experiment,
such as taking a single glimpse. For the case where a sequence of glimpses can
be taken, we need \textit{sequential} experimental design. In this scenario, the
choice of design $l_t$ can be informed by the designs and outcomes of previous
experiments, $l_{1:t-1}$ and $\y_{1:t-1}$.
The marginal distribution over outcomes is therefore $p(\y_t|l_{1:t},\y_{1:t-1})$
rather than $p(\y_t|l_t)$.
Similarly, the posterior after observing $\y_t$ is $p(\theta|l_{1:t},\y_{1:t})$.
Therefore, in the sequential case which we consider throughout the rest of the
paper, we minimise the following form of the $\EPE$:
\begin{align}
  \label{eq:seq-epe}
  \text{EPE}_{\y_{1:t-1}, l_{1:t-1}} (l_t) = \EX_{p(\y_t | \y_{1:t-1}, l_{1:t})} \left[\entropy \left[ p(\theta | \y_{1:t}, l_{1:t} ) \right]  \right].
\end{align}
To summarise, sequential BOED involves, at each time $t$, selecting $l_t =
\argmin_{l_t} \text{EPE}_{\y_{1:t-1}, l_{1:t-1}} (l_t)$ and then performing the
experiment with design $l_t$ to observe $\y_t$.

\section{Generating near-optimal glimpse sequences} \label{sec:nogs}

\ourparagraph{Role of BOED pipeline}
To reiterate the outline of our method,
we first annotate a portion of the training data with glimpse sequences, and
then in the second stage use these to speed up the training of a hard attention
mechanism. This section details our BOED pipeline for the first stage.

\ourparagraph{EPE estimator}
BOED requires a probabilistic model of the measurements and parameters we wish
to infer. That is, we need to define $p(\theta, \y_{1:t} | l_{1:t})$ for any
$l_{1:t}$. To do so in the visual attention setting, we first define $p(\theta,
\x)$ to be the intractable joint distribution over labels and images from which
our training and test data originate. %
To be consistent with our definition in \cref{sec:hard-attention} of $\y$ as a
deterministic function of $\x$ and $l$, we then define $p(\y_i|\x,l_i)$ to be a
Dirac-delta distribution on $\fovea(\x,l_i)$.
The joint distribution is then
\begin{equation} \label{eq:p-joint}
  p(\theta, \y_{1:t}|l_{1:t}) = \int p(\theta, \x) \prod_{i=1}^t p(\y_i|\x,l_i) \mathrm{d}\x.
\end{equation}
Given this joint distribution, $\text{EPE}_{\y_{1:t-1}, l_{1:t-1}} (l_t)$ is
well defined but intractable in general. We therefore consider how to
approximate it. To simplify our method for doing do, we first rearrange the
expression given in \lref{eq:seq-epe} so that the expectation is over $\x$
rather than $\y_t$. Taking advantage of the fact that $\y_i$ is a deterministic
function of $\x$ and $l_i$ allows it to be rewritten as follows (proof in the
appendix). Defining $\fovea(\x, l_{1:t}) = \{ \fovea(\x,
l_{1}),\ldots,\fovea(\x, l_t) \}$,
\begin{equation} \label{eq:image-epe}
  \text{EPE}_{\y_{1:t-1}, l_{1:t-1}} (l_t) = \EX_{p(\x | \y_{1:t-1}, l_{1:t-1})} \left[\entropy \left[ p(\theta | \fovea(\x, l_{1:t}), l_{1:t} ) \right]  \right].
\end{equation}
Given this form of the expected posterior entropy, we can approximate it if we
can leverage the dataset to make the following two approximations:
\begin{itemize}[itemsep=0pt, topsep=0pt]
\item a learned \textit{attentional variational posterior}, $\gavp(\theta | \y_{1:t}, l_{1:t} )
  \approx p(\theta | \y_{1:t}, l_{1:t} )$,
\item and \textit{stochastic image completion} distribution
  $\rimg(\x|\y_{1:t-1},l_{1:t-1}) \approx p(\x|\y_{1:t-1},l_{1:t-1})$.
\end{itemize}
We expand on the form of each of these approximations later in this section.
First, combining them with \cref{eq:image-epe} and using a Monte Carlo estimate
of the expectation yields our estimator for the $\EPE$:
\begin{align} \label{eq:approx-epe}
  \text{EPE}_{\y_{1:t-1}, l_{1:t-1}} \left( l_{t} \right) \approx \frac{1}{N} \sum_{n=1}^N \entropy \big[ \gavp(\theta | \fovea(\x^{(n)}, l_{1:t}), l_{1:t} ) \big]
\end{align}
with $\x^{(1)}\ldots,\x^{(N)} \sim \rimg(\x|\y_{1:t-1},l_{1:t-1})$.

\ourparagraph{Overview of BOED pipeline}
We select $l_t$ with a grid search. That is, denoting the set of allowed values
of $l_t$ as $L$, we compute our approximation of $\text{EPE}_{\y_{1:t-1},
  l_{1:t-1}} (l_t)$ for all $l_t \in L$. We then select the value of $l_t$ for
which this is least. To do so, our full BOED pipeline is as follows.
\begin{enumerate}[itemsep=0pt, topsep=0pt]
\item Sample $\x^{(1)}\ldots,\x^{(N)} \sim \rimg(\x|\y_{1:t-1},l_{1:t-1})$.
\item For each $l_t \in L$, approximate the expected posterior entropy with
  \lref{eq:approx-epe}.
\item Select the value of $l_t$ for which this approximation is least.
\end{enumerate}
This is repeated for each $t = 1,\ldots,T$. \Cref{fig:epe-maps} shows examples
of $\EPE$s estimated in this way for each $t$. We now detail the form of
$\gavp$ (the attentional variational posterior) and $\rimg$ (stochastic
image completion).

\begin{figure}
  \centering
  \includegraphics[scale=1]{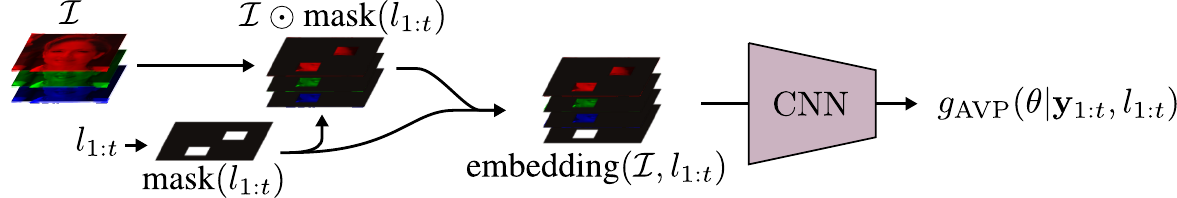}
  \caption{Attentional variational posterior CNN.
    An RGB image and $l_{1:t}$ are processed to create an
    embedding of the information gained from glimpses $1$ to $t$.
    This embedding is fed into an image classifier to obtain an approximation of
    $p(\theta|\y_{1:t}, l_{1:t})$.}
  \label{fig:dropout-cnn}
\end{figure}





\ourparagraph{Attentional variational posterior} In this section we introduce
our novel approach to efficiently approximating the intractable posterior
$p(\theta | \y_{1:t}, l_{1:t})$. We train a convolutional neural network (CNN)
to map from a sequence of glimpses , $\y_{1:t}$, and their locations, $l_{1:t}$,
to $\gavp(\theta | \y_{1:t}, l_{1:t})$, an approximation of this posterior. We
call this the attentional variational posterior CNN (AVP-CNN). To allow a single
CNN to cope with varying $\y_{1:t}$, $l_{1:t}$, and even varying $t$, we embed
its input as shown in \lref{fig:dropout-cnn}. Essentially, $l_{1:t}$ is used to
create a mask the size of the image which is 1 for observed pixels and 0 for
unobserved pixels. Elementwise multiplication of this mask with the input image
sets unobserved pixels to zero. The mask is then concatenated as an additional
channel. This embedding naturally maintains spatial information while enforcing
an invariance to permutations of the glimpse sequence. Our experiments use a
Densenet-121~\cite{huang2017densely} CNN architecture (pretrained on
ImageNet~\cite{imagenet}) to map from this embedding to a vector of class
probabilities representing $\gavp$.


We train the network to minimise the KL divergence between its output and
$p(\theta|\y_{1:t},l_{1:t})$. That is, $D_{KL} \left( p(\theta|\y_{1:t},l_{1:t})
  || \gavp(\theta|\y_{1:t},l_{1:t}) \right) $. To ensure that $\gavp$ is close
for all $t$, $l_{1:t}$ and $\y_{1:t}$, the loss used is an expectation of this
KL divergence over $p(\y_{1:t}|l_{1:t})u(t, l_{1:t})$. We factorise $u(t,
l_{1:t})$ as $u(t)\prod_{i=1}^tu(l_i)$ where, so that all times and glimpse
locations are weighted equally in the loss, $u(t)$ is a uniform distribution
over $1,\ldots,T$ and $u(l_i)$ is a uniform distribution over all image
locations. Denoting the network parameters $\lambda$, the gradient of this loss
is
\begin{align} \label{eq:dropout-cnn-gradient}
  \frac{\partial}{\partial\lambda} \mathcal{L}_\lambda = \EX_{p(\theta, \y_{1:t}|l_{1:t})u(t, l_{1:t})} \left[ - \frac{\partial}{\partial\lambda} \log g_\text{AVP}^{\lambda}(\theta | \y_{1:t}, l_{1:t}) \right].
\end{align}
This gradient is the same as that of a cross-entropy loss on data sampled from
$p(\theta, \y_{1:t}|l_{1:t})u(t, l_{1:t})$, and can be approximated by a Monte
Carlo estimate.

Our approximation of the $\EPE$ in \cref{eq:approx-epe} involves the entropy of
$\gavp$. Since $\gavp$ is a categorical distribution, this is simple to compute
analytically. Our approximation of the posterior entropy by an amortised
artifact in this way is inspired by the variational posterior estimator
introduced by \citet{foster2019variational}, although there are two important
differences:
\begin{itemize}[itemsep=0pt, topsep=0pt]
\item \citeauthor{foster2019variational} learn a mapping from $\y_t$ to
  $g(\theta | \y_{1:t}, l_{1:t})$, sharing information between ``nearby''
  samples of $\y_t$ to reduce the computational cost of the experimental design.
  Our AVP-CNN takes this amortization further by learning a single mapping from
  $t$, $l_{1:t}$ and $\y_{1:t}$ to $\gavp(\theta | \y_{1:t}, l_{1:t})$, which
  yields significant further efficiency gains in our setting.
\item Whereas we approximate $\entropy[p]$ with $\entropy[\gavp] = \EX_\gavp[-\log\gavp]$,
  \citeauthor{foster2019variational} use $\EX_p[-\log g]$. This provides an
  upper bound on $\entropy[p]$ but is not applicable in our case as we cannot
  sample from $p(\theta|\y_{1:t}, l_{1:t})$. Both approximations are exact when
  $\gavp = p$.
\end{itemize}

\ourparagraph{Stochastic image completion}
\label{sec:image-sampling}
We considered numerous ways to form $\rimg(\x | \y_{1:t-1}, l_{1:t-1})$
including inpainting techniques~\cite{pathak2016context,isola2017image} and
through Markov chain Monte Carlo in a probabilistic image model. Future research
in GANs and generative modelling may provide better alternatives to this
component of our method but, for now, we choose to represent $\rimg$ using a
technique we developed based on image retrieval~\cite{jegou2010improving}. We
found that, of the methods we considered, this gives the best trade-off between
speed and sample quality. It involved creating an empirical image distribution
consisting of 1.5 million images for each experiment using deep generative
models with publicly available pre-trained weights
(StyleGAN~\cite{karras2018style} for CelebA-HQ and
FineGAN~\cite{singh2019finegan} for Caltech-UCSD Birds). During sampling, the
database is searched for images that `match' the previous glimpses ($\y_{1:t-1}$
and $l_{1:t-1}$). How well these glimpses match some image in the database,
$\x'$, is measured by the squared distance in pixel space at glimpse locations:
$\sum_{i=1}^{t-1} \norm{\y_i - \fovea(\x', l_i)}^2_2$. This distance is used to
define a probability distribution over the images in the database.
To reduce computation, we first cheaply compare approximations of the observed
parts of each image using principal component
analysis~\cite{jolliffe2011principal}, and compute exact distances only when
these are close. The overall procedure to sample from $\rimg$ corresponds to
importance sampling~\cite{arulampalam2002tutorial} in a probabilistic model
where $p(\y_t | \x, l_t)$ is relaxed from a Dirac-delta distribution to a
Gaussian. See the appendix for further details.


\section{Training with partial supervision} \label{sec:sup}
\label{sec:partially-supervised-training}
The previous section describes how to annotate an image with a near-optimal
sequence of glimpse locations for a particular image classification task. This
section assumes that these, or other forms of glimpse sequence (e.g. the
handcrafted glimpse sequences in \lref{sec:experiments}), exist for all, or
some, images in a dataset. These can then be used to partially supervise the
training of a hard attention mechanism on this dataset. We refer to glimpse
sequences used in this way as supervision sequences. We use separate losses for
supervised (i.e. annotated with both a class label and a sequence of glimpse
locations) and unsupervised (i.e. annotated with a class label but not glimpse
locations) examples. By minimising the sum of these losses, our procedure can be
viewed as maximising the joint log-likelihood of the class labels and
supervision sequences.
To be precise, let $q_\phi (\theta^i, l_{1:T}^i | \x^i)$ be a network's joint
distribution over the chosen glimpse locations and predicted class label on
image $\x^i$. Let $q_\phi (\theta^i | \x^i)$ be the marginalisation of this
distribution over $l^i_{1:t}$. We maximise a lower bound on
\begin{equation}
  \label{eq:objective}
  \mathbf{L} = \sum_{i \in \text{sup.}} \overbrace{\log q_\phi (\theta^i, l_{1:T}^i | \x^i)}^{\text{supervised objective}} + \sum_{i \in \text{unsup.}} \overbrace{\log q_\phi (\theta^i | \x^i)}^{\text{unsupervised objective}}.
\end{equation}
where `sup' is the set of training indices with supervision sequences, and
`unsup' is the remainder.

When running on unsupervised examples, we follow \citet{mnih2014recurrent} and
train the location network with a REINFORCE estimate of the gradient of the
accuracy, using a learned baseline to reduce the variance of this estimate.
Meanwhile, the RNN, glimpse embedder, and classifier network are trained to
maximise the log-likelihood of the class labels (i.e. minimise a cross-entropy
loss). \citet{ba2014multiple} noted that this can be viewed as maximising a
lower bound on the unsupervised objective in \lref{eq:objective}.
For examples with supervision sequences, the supervised objective in
\lref{eq:objective} is maximised by gradient backpropagation. The loss is
computed by running the network with its glimpse locations fixed to those in the
supervision sequence. The location network is updated to maximise the
probability of outputting these glimpse locations while, as for unsupervised
examples, the other network modules are trained to maximise the likelihood of
the class labels. Gradients of the supervised and unsupervised objectives can be
computed simultaneously, with minibatches containing both types of example.
 
\begin{figure}[t]
  \centering
  \includegraphics[scale=1]{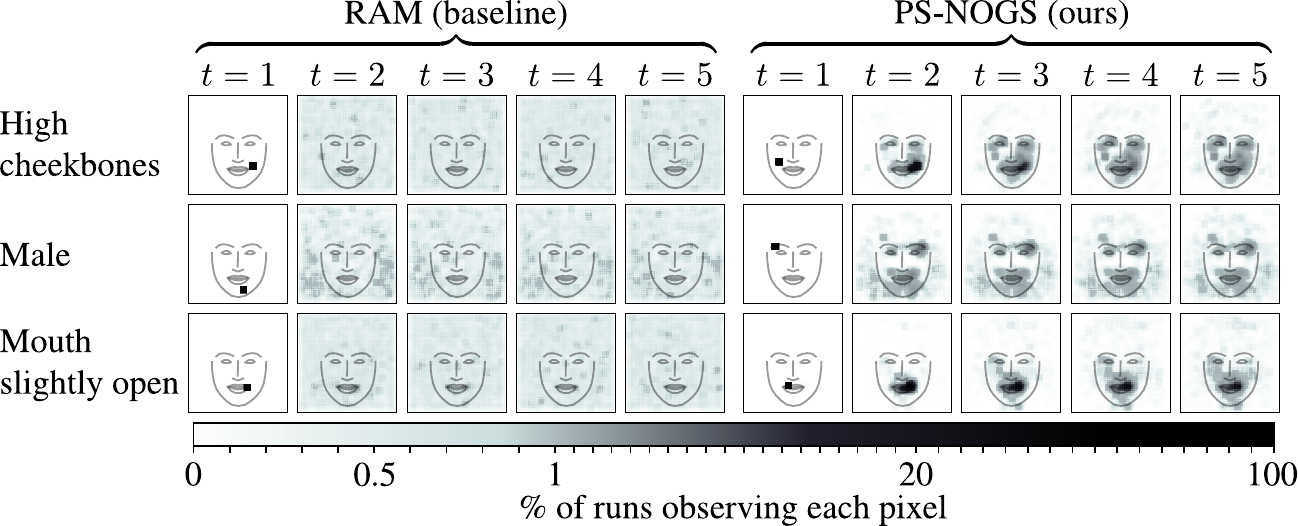}
  \caption{Comparison of glimpse locations chosen by \RAM{} and \PSNOGS{} on the
    CelebA-HQ test set for three classification tasks. For each $t \in
    \{1,2,3,4,5\}$, we show an image where each pixel's colour corresponds to
    how often it was observed at this time step during testing. The outlines are
    produced by averaging outputs from a face detector across the dataset. %
    For $t=1$, each network learns a single location which it attends to on
    every test image. This is expected behaviour as the first location is chosen
    before taking any glimpses, and therefore before being able to condition on
    the image.
    \RAM{} appears to then fail to learn to direct the later glimpses, attending
    almost uniformly across the image.
    In contrast, \PSNOGS{} distributes these glimpses broadly over the salient
    regions. }
  \vspace{-.3cm}
  \label{fig:glimpses}
\end{figure}

\section{Experiments and results} \label{sec:exp}
\label{sec:experiments}
\ourparagraph{Datasets and network architectures}
We test our approach on CelebA-HQ~\cite{karras2017progressive} and a cropped
variant of Caltech-UCSD Birds (CUB)~\cite{wah2011caltech}, both of which are
convincingly modelled by state-of-the-art
GANs~\cite{karras2018style,singh2019finegan} as we require.
For both datasets, we use $T=5$. The hard attention network's classifier is a
fully-connected layer mapping from the hidden state to a softmax output and the
location network has a single 32-dimensional hidden layer. All hard attention
networks are trained by Adam optimiser~\cite{kingma2014adam} with a batch size
of 64; learning rate $3\times10^{-4}$ for CelebA-HQ and $5\times10^{-5}$ for
CUB; and other hyperparameters set to the recommended
defaults~\cite{kingma2014adam}. The dataset-specific details are as follows:
\textbf{(1) CelebA-HQ} Our experiments tackle 40 different binary classification
tasks, corresponding to the 40 labelled attributes. We resize the images to $224
\times 224$ and use training, validation, and test sets of $27\,000$, $500$, and
$2500$ images respectively. We use $16\times16$ pixel glimpses, with a
$50\times50$ grid of allowed glimpse locations. The glimpse network has two
convolutional layers followed by a linear layer, and the RNN is a
GRU~\cite{cho2014learning} with hidden dimension $64$. \textbf{(2) CUB} We
perform 200-way classification of bird species. We crop the images using the
provided bounding boxes and resize them to $128 \times 128$. Cropping is
necessary because good generative models do not exist for the uncropped dataset,
but there is still considerable variation in pose after cropping. We use 5120
training images, a validation set of 874 and a test set of 5751 (having removed
43 images also found in ImageNet). We use $32\times32$ pixel glimpses and a
$12\times12$ grid of allowed glimpse locations so that adjacent locations are 8
pixels apart. The glimpse network is the first 12 convolutional layers of a VGG
pretrained on ImageNet~\cite{simonyan2014very, imagenet}, and the RNN is a GRU
with hidden dimension $1024$.

\begin{figure*}
  \centering
  \includegraphics[scale=1]{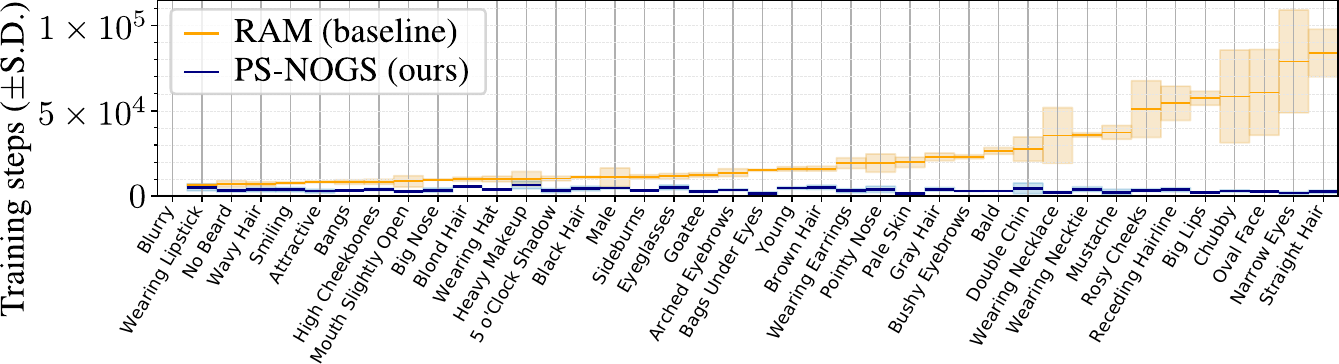}
  \caption{Number of training iterations for each CelebA-HQ attribute before a
    validation cross-entropy loss within 0.01 of the best is achieved. On
    average, \PSNOGS{} trains almost $7\times$ faster than \RAM{} and with less than a
    fifth of the variance in training speed. This speed-up is complemented by an
    average increase in test accuracy of $0.4\%$. Attributes are sorted by \RAM's
    mean training time.}
  \label{fig:celebhq-time}
\end{figure*}

\ourparagraph{BOED} We create $600$ near-optimal glimpse sequences for each of
the 40 CelebA-HQ classification tasks, and $1000$ for CUB. This took
approximately 20 GPU-hours for CUB, and 10 GPU days for each CelebA-HQ
attribute. We have publicly released these sequences along with our
code\footnote{Our code and near-optimal glimpse sequences are at
  \url{https://github.com/wsgharvey/ps-nogs}.}, allowing them to be re-used by
anyone to speed up the training of hard attention networks on these tasks.

\ourparagraph{RAM baseline} On both experiments, we train our architecture with
the algorithm used for the recurrent attention model (\RAM) of
\citet{mnih2014recurrent} as a baseline, which is equivalent to the special case
of our partially supervised objective with zero supervision sequences. We
compare this to our method of partially-supervising training using near-optimal
glimpse sequences (\PSNOGS{}).

\ourparagraph{Partial-supervision for CelebA-HQ} In \lref{fig:celebhq-time}, we
plot the number of iterations until convergence for RAM and \PSNOGS{} on
each different CelebA-HQ classification task. Using \PSNOGS{} reduced the
average number of iterations by a factor of 6.8 compared to \RAM{} ($3500$ vs.
$24\,000$) while improving mean test accuracy from $86.7\%$ to $87.1\%$. We also
compare the attention policies learned by each method. These are plotted for
several tasks in \lref{fig:glimpses}, and for the remainder in the appendix.
\RAM{} has difficulty learning policies for more than the first glimpse.
\PSNOGS{} appears to learn a reasonable policy for every timestep.

\begin{wrapfigure}{r}{0.5\textwidth}
  \centering
  \vspace{-.25cm}
  \includegraphics[scale=1]{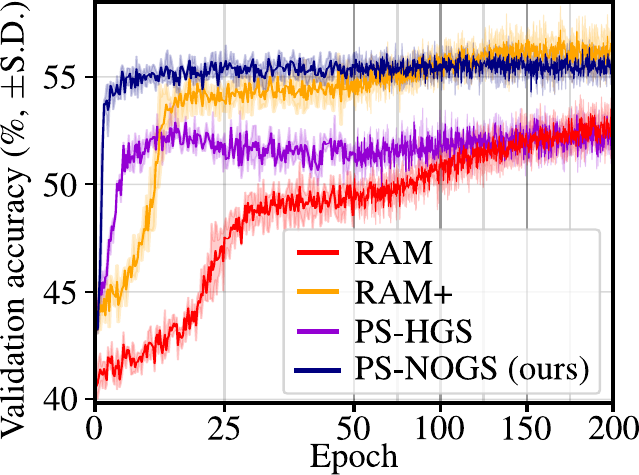}
  \caption{CUB validation accuracy over training. }
  \vspace{-.4cm}
  \label{fig:birds-training}
\end{wrapfigure}
\ourparagraph{Partial-supervision for CUB} For experiments on CUB, we find that
a pretraining stage is required to obtain good final classification accuracy. We
therefore pretrain the classifier, RNN, and glimpse network with glimpse
locations sampled independently at each time step from either a uniform
distribution (in the case of \RAM) or from a heuristic which assigns
higher probability to more salient locations, as estimated using the AVP-CNN
(\RAMP{} and all others). Additional details are available in the appendix. We
train both \RAM{} baselines and \PSNOGS{}. We also consider partial supervision
with hand-crafted glimpse sequences, \HGS{}. The hand-crafted glimpse sequences
are designed, using CUB's hand-annotated features, to always attend to the beak,
eye, forehead, belly and feet (in that order). If any of these parts are
obscured, they are replaced by a randomly selected visible body part.

\begin{table}
  \caption{Results summary }
  \label{tab:results}
  \centering
  \begin{tabular}{lllll}
    \toprule
    \multicolumn{1}{r}{} & \multicolumn{2}{c}{CelebA-HQ (avg.)} & \multicolumn{2}{c}{CUB}                 \\
    \cmidrule(r){2-3} \cmidrule(r){4-5}
    Method     & Iterations     & Accuracy (\%)  & Iterations   & Accuracy (\%) \\
    \midrule
    \RAM{} / \RAMP{} & 24\,000  & 86.7 & 11\,600 & \textbf{56.3}     \\
    \HGS{}       & - & -  & 960  & 52.0     \\
    \PSNOGS{} (ours)     & \textbf{3500}       & \textbf{87.1}  & \textbf{640}  & 55.4 \\
    \bottomrule
  \end{tabular}
\end{table}

In \Cref{fig:birds-training}, we plot the validation accuracy of the various
methods throughout training. We find that \PSNOGS{} outperforms the baselines at
the start of training, achieving $54\%$ accuracy after just 5 epochs. In
comparison, the \RAMP{} model takes 16 epochs to reach $54\%$ and \HGS{} never
does. The naive baseline \RAM{} has worse validation accuracy throughout
training than \RAMP{}, which uses the AVP-CNN for pre-training. Finally, we note
that after ~95 epochs, the validation accuracy for \RAMP{} overtakes that for
\PSNOGS{}. This may be because REINFORCE has unbiased gradient estimates,
whereas error due to approximations in the generation of supervision sequences
leads to a bias throughout training. \Cref{tab:results} summarises the accuracy
and iterations until convergence on each dataset, with the best technique on
each metric in bold. The number of iterations on CUB is the number before
achieving a validation accuracy within 1\% of the highest.

\section{Related work} \label{sec:related-work}

\ourparagraph{Hard attention architectures}
\citet{elsayed2019saccader} recently demonstrated a hard attention network which
achieved accuracy on ImageNet~\cite{imagenet} close to that of CNNs which use
the whole image. However, their approach neccesitates running a convolutional
network on the entire image to select glimpse locations. As such, they advertise
improvements in interpretability rather than computational efficiency.
\citet{sermanet2014attention} train a hard attention architecture with REINFORCE
to achieve state-of-the-art accuracy on the Stanford Dogs dataset. In addition
to accessing the full image in low resolution at the start, they use large
glimpses (multiple $96\times96$ pixel patches at different resolutions) to
effectively solve the task in a single step. This avoids problems resulting from
learning long sequences with REINFORCE but also rules out the computational
gains possible with smaller glimpses. \citet{katharopoulos2019processing}
proposed a form of hard attention where, after processing the downsampled image,
multiple glimpses are sampled and processed simultaneuosly. This is again
incompatible with a low-power setting where we cannot afford to operate on the
full image.




\ourparagraph{Supervised attention} One solution to improving training is to
provide supervision targets for the attention mechanism. This is common in
visual question answering, usually for training soft attention. Typically the
targets are created by human subjects, either with
gaze-tracking~\cite{yu2017supervising} or explicit data
annotation~\cite{das2017human}. Either way is expensive and
dataset-specific~\cite{das2017human, gan2017vqs}. Recent work has considered
reducing this cost by, for example, extrapolating supervision signals from a
manually-annotated dataset to other, related, datasets~\cite{qiao2018exploring},
or using existing segmentations to speed up annotation~\cite{gan2017vqs}. Even
so, considerable human effort is required. Our method for generating
near-optimal glimpse sequences can be viewed as automation of this in an image
classification context.

\section{Discussion and conclusion}

We have demonstrated a novel BOED pipeline for generating near-optimal
sequences of glimpse locations.
We also introduced a partially supervised training objective which uses such a
supervision signal to speed up the training of a hard attention mechanism.
By investing up-front computation in creating near-optimal glimpse sequences for
supervision, this speed up can be achieved along with comparable final accuracy.
Since we release the near-optimal glimpse sequences we generated, faster
training and experimentation on these tasks is available to the public without
the cost of generating new sequences.
Our work could also have applications in neural architecture search, where this
cost can be amortised over many training runs with different architectures.

There is potential to significantly speed up the BOED pipeline by, for example,
selecting $l_t$ though Bayesian optimisation rather than a grid search. Our
framework could also be extended to attention tasks such as question answering
where the latent variable of interest is richly structured.

\section*{Broader Impact}
Our work on hard attention aims to make possible more power-efficient computer
vision. This could make certain computer vision methods more feasible in
settings where limited power is available, such as in embedded devices, or for
applications with extremely large data throughput, such as processing satellite
images. These applications can have a positive impact through, for example,
environmental monitoring or healthcare but also potential negative implications,
particularly with regard to privacy. We hope that standards and legislation can
be developed to mitigate these concerns.

Another motivation for hard visual attention is its
interpretability~\cite{elsayed2019saccader,xu2015show}, since a network's
outputs can be more easily understood if we know which part of the image
informed them. This is especially applicable to the hard attention architecture
considered in our work, which never accesses the full image. Greater traction
for such architectures, whether motivated by power efficiency or
interpretability, would therefore lead to computer vision systems making more
interpretable decisions. This may lead to more trust being placed in such
systems. An important research direction is the implications of this trust, and
whether it is justified by this level of interpretability.

\begin{ack}
  We acknowledge the support of the Natural Sciences and Engineering Research
  Council of Canada (NSERC), the Canada CIFAR AI Chairs Program, and the Intel
  Parallel Computing Centers program. This material is based upon work supported
  by the United States Air Force Research Laboratory (AFRL) under the Defense
  Advanced Research Projects Agency (DARPA) Data Driven Discovery Models (D3M)
  program (Contract No. FA8750-19-2-0222) and Learning with Less Labels (LwLL)
  program (Contract No.FA8750-19-C-0515). Additional support was provided by
  UBC's Composites Research Network (CRN), Data Science Institute (DSI) and
  Support for Teams to Advance Interdisciplinary Research (STAIR) Grants. This
  research was enabled in part by technical support and computational resources
  provided by WestGrid (\url{https://www.westgrid.ca}) and Compute Canada
  (\url{www.computecanada.ca}). William Harvey acknowledges support by the
  University of British Columbia's Four Year Doctoral Fellowship (4YF) program.
\end{ack}

\bibliography{bibfile}
\bibliographystyle{unsrtnat}

\clearpage

\appendix







\section{Details for method}
\subsection{BOED pseudocode}
\begin{figure}[h]
  \centering
  \includegraphics[scale=1]{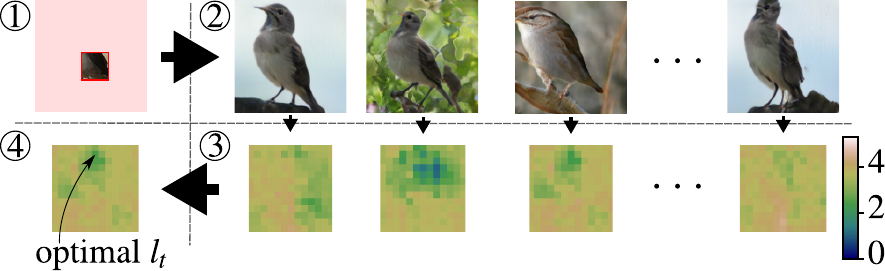}
  \caption{ BOED procedure to select $l_2$ given $\y_1$ and $l_1$, corresponding
    to \cref{alg:boed} with $t=2$. Panel 1 shows a visualisation of the single
    previous glimpse, $\y_1$, taken at $l_1$, with all unobserved image pixels
    displayed in pink. %
    Panel 2 shows $\x^{(1)},\ldots,\x^{(N)}$, Monte Carlo samples of the full
    image drawn on line~\ref{line:image-sampling} of \cref{alg:boed}. %
    For each of these images, panel 3 shows a heatmap. Each heatmap plots the
    estimated posterior entropy after conditioning on $\y_t = f_\text{fovea}(\x,
    l^*)$ and $l_t=l^*$ for each $l^* \in L$, plotted against the $x$ and $y$
    coordinates of $l^*$. These posterior entropies are estimated on
    line~\ref{line:pe}, and then averaged over the Monte Carlo samples on
    line~\ref{line:epe}, giving the \textit{expected} posterior entropies
    displayed by a heatmap in panel 4. Finally, $l_t$ is chosen to minimise
    $\EPE_{l_t}$. } \label{fig:boed-overview}
\end{figure}

\Cref{alg:boed} gives pseudocode for our BOED pipeline. This selects a near-optimal
glimpse location for a single timestep. Repeating this for $t=1,\ldots,T$ yields
a near-optimal glimpse sequence for a particular image. \Cref{fig:boed-overview}
provides a visualisation of the various stages of this algorithm.
\begin{algorithm}[H]
  \begin{algorithmic}[1]
    \Procedure{SelectLocation}{$\y_{1:t-1}, l_{1:t-1}$}
    \State $\x^{(1)},\ldots,\x^{(N)} \sim r_{\text{img}}(\x | \y_{1:t-1}, l_{1:t-1})$  \Comment{stochastic image completion} \label{line:image-sampling}
    \For{$l_t \in L$} \Comment{grid search for $l_t$} \label{line:grid-search}
    \For{$n \gets 1,\ldots,N$}
    \State $\text{PE}_{l_t}^{(n)} \gets \entropy\left[ \gavp(\theta |
      \fovea(\x^{(n)}, l_{1:t}), l_{1:t} ) \right]$ \label{line:pe}
    \Comment{estimate posterior entropy}
    \EndFor
    \State $\EPE_{l_t} \gets \frac{1}{N} \sum_{n=1}^N
    \text{PE}_{l_t}^{(n)}$ \label{line:epe}  \Comment{average over Monte Carlo samples}
    \EndFor
    \State $l_t^* \gets \mathrm{argmin}_{l_t} \EPE_{l_t}$  \label{line:min}
    \State \textbf{return} $l_t^*$
    \EndProcedure
  \end{algorithmic}
  \caption{BOED pipeline to select $l_t$. This roughly follows the three-step
    procedure given in \lref{sec:nogs}: $N$ images are sampled in
    line~\ref{line:image-sampling}; these are
    used to estimate expected posterior entropies for each $l_t \in L$ in
    lines~\ref{line:grid-search}-\ref{line:epe}. Finally, the minimising value
    of $l_t$ is found on line~\ref{line:min}.}.
  \label{alg:boed}
\end{algorithm}

\subsection{EIG estimator in \lref{eq:image-epe}}
This section derives the form of the expected posterior entropy presented in
\lref{eq:image-epe}. Starting from \lref{eq:seq-epe}, we introduce an inner
expectation over $\x$:
\begin{align}
  \text{EPE}_{\y_{1:t-1}, l_{1:t-1}} (l_t) &= \EX_{p(\y_t | \y_{1:t-1}, l_{1:t})} \left[\entropy \left[ p(\theta | \y_{1:t}, l_{1:t} ) \right]  \right] \\
                                           &= \EX_{p(\y_t | \y_{1:t-1}, l_{1:t})} \EX_{p(\x|\y_{1:t},l_{1:t})} \left[\entropy \left[ p(\theta | \y_{1:t}, l_{1:t} ) \right]  \right] \\
                                           &= \EX_{p(\y_t, \x | \y_{1:t-1}, l_{1:t})} \left[\entropy \left[ p(\theta | \y_{1:t}, l_{1:t} ) \right]  \right].
\end{align}
Since, according to the probabilistic model defined in \lref{eq:p-joint}, each
$\y_i$ is a deterministic function of $\x$ and $l_i$ we can substitute
$\y_{1:t}$ using the definition $\y_{1:t} = \fovea(\x, l_{1:t}) = \{ \fovea(\x,
l_{1}),\ldots,\fovea(\x, l_t) \}$.
\begin{align}
  \text{EPE}_{\y_{1:t-1}, l_{1:t-1}} (l_t) &= \EX_{p(\y_t, \x | \y_{1:t-1}, l_{1:t})} \left[\entropy \left[ p(\theta | \fovea(\x, l_{1:t}), l_{1:t} ) \right]  \right] \\
                                           &= \EX_{p(\x | \y_{1:t-1}, l_{1:t-1})} \left[\entropy \left[ p(\theta | \fovea(\x, l_{1:t}), l_{1:t} ) \right]  \right]
\end{align}
as presented in \lref{eq:image-epe}.

\subsection{Stochastic image completion}
This section provides more detail on our method for stochastic image completion
and the alternatives we considered, as discussed in \cref{sec:nogs}. Three
sampling mechanisms for $\rimg(\x | \y_{1:t-1}, l_{1:t-1})$ were compared
qualitatively, based on both the diversity of samples produced and how realistic
the samples are (taking into account the previous observations). These were a
conditional GAN, HMC in a deep generative model, and the image retrieval-based
approach used in our pipeline. We now provide more detail on each.

\subsubsection{Conditional GAN}
We considered a conditional GAN, motivated by recent state-of-the-art
performance on conditional image generation
problems~\cite{nazeri2019edgeconnect, pathak2016context}. In particular, we
considered using pix2pix~\cite{isola2017image} to map from an embedding of
$\y_{1:t}$ and $l_{1:t}$ (similar to that in \lref{fig:dropout-cnn} but without
concatenating the mask, and with unobserved pixels replaced with Gaussian noise
instead of zeros) to the completed image. We used the U-net
architecture~\cite{ronneberger2015u} with resolution $256\times256$.
\lref{fig:image-sampling-comparison} shows samples when all other
hyperparameters were set to the defaults~\cite{isola2017image}. We tried varying
both the generator architecture and the weight given to the L1-norm, although
neither significantly improved the quality of the generated images. Although the
GAN can learn to generate realistic images for CelebA-HQ, they have very little
diversity when the observations are fixed. An additional issue is that the
outputs looked considerably less realistic for the Caltech-UCSD birds dataset.

\subsubsection{HMC in a deep generative model}
\begin{figure}
  \centering
  \includegraphics[width=0.7\textwidth]{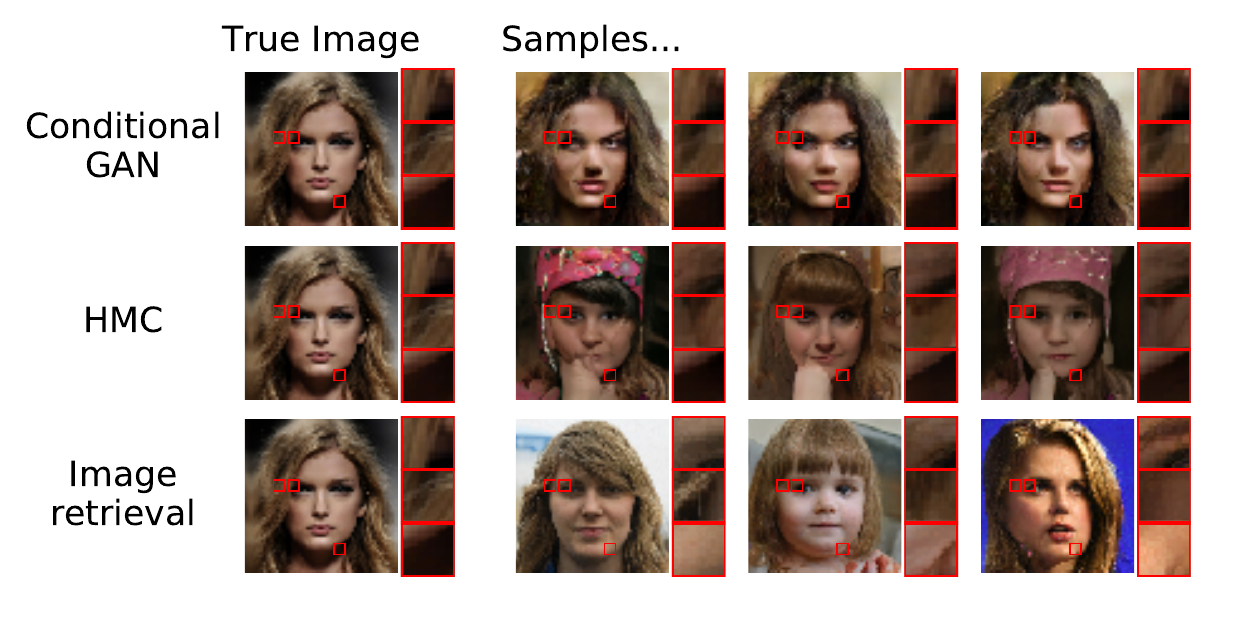}
  \caption{Sampled image completions using various techniques. The glimpse
    locations are marked with red squares, and close-ups of each glimpse are
    shown to the right of each image. The leftmost column shows the `true' image
    from which the observed glimpses are taken and the others show samples
    conditioned on these glimpses.}
  \label{fig:image-sampling-comparison}
\end{figure}

We also considered taking a pre-existing, unconditional, GAN
(StyleGAN~\cite{karras2018style}) with publicly available weights and using
Hamiltonian Monte Carlo~\cite{duane1987hybrid, hoffman2014no} (HMC) to sample
images from it conditioned on the observations. Specifically, we sample from
$p(\x|\y_{1:t}, l_{1:t}) \propto p(\x)\prod_{i=1}^t p(\y_i | \x, l_i)$, where
$p(\x)$ is the prior over images defined by the GAN. Deviating from the model
specified in \cref{eq:p-joint}, we define $p(\y_i|\x, l_i)$ to be an isotropic
Gaussian centred on $\fovea(\x, l_i)$, rather than a Dirac-delta on $\fovea(\x,
l_i)$. This change is necessary to make inference with HMC feasible. We used an
implementation of HMC in Pyro~\cite{bingham2018pyro}, a probabilistic
programming language. Samples are shown in the second row of
\cref{fig:image-sampling-comparison}. These were taken using a likelihood
with standard deviation 0.4 applied to the normalised pixel values, which was
set to trade-off the `closeness' of samples to the true image and the
feasibility of inference. The integration was performed with 200 steps of size
0.05 and all other parameters set to the defaults. Although the samples mostly
look realistic, they suffered from low sample diversity; the HMC chains became
stuck in certain modes and did not explore the full posterior. Additionally,
sampling was slow: drawing each sample took approximately a minute on a GPU.

\subsubsection{Image retrieval} \label{ap:image-retrieval}

\begin{algorithm*}[t]
  \caption{\textsc{SampleImages} calls \textsc{SampleProposal} to select $K_1$
    images to load into memory, based on approximations of their importance
    weights. These are then reweighted using their exact likelihood. $K_2$
    images are sampled from the resulting weighted categorical distribution, and
    returned.}
  \label{alg:image-sampling}
  \begin{algorithmic}
    \Procedure{SampleProposal}{$\x, l_{1:t}$}
    \State Select relevant weight matrix columns $\mathbf{\tilde{W}} \gets \textsc{Slice}(\mathbf{W}, l_{1:t}) $
    \State Select relevant columns of mean vector $\mathbf{\tilde{\mu}} \gets \textsc{Slice}(\mathbf{\mu}, l_{1:t}) $
    \State Reconstruct observations $\mathbf{\hat{y}}_{1:t} \gets
    \mathbf{\tilde{\mu}} + \mathbf{\tilde{W}}^\top\mathbf{W} \x $
    \For{$i = 1,\ldots,\ndata{}$}
    \State Approximate observed patches $\mathbf{\hat{y}}_{1:t}^i \gets \mathbf{\tilde{W}}^\top \mathbf{z}^i $
    \State Compute approximate likelihood $w^i_1 \gets \mathcal{N}\left(
      \mathbf{\hat{y}}_{1:t} | \mathbf{\hat{y}}_{1:t}^i, \sigma_q^2 \right)$
    \EndFor
    \State $j^{(1)}, \ldots, j^{(K_1)} \gets \textsc{Resample}(w^1_1, \ldots, w_1^\ndata{}) $
    \State \textbf{return} $\{i^{(k)}, w_1^{i^{(k)}}\} \text{ for }
    k=1,\ldots,K_1$
    \EndProcedure
  \end{algorithmic}
  \begin{algorithmic}
\Procedure{SampleImages}{$\x, l_{1:t}$}
    \State {\bfseries Input:} $\x, l_{1:t}$
    \State $\{i^{(1)}, w^{i^{(1)}}_1\}, \ldots, \{i^{(K_1)}, w^{i^{(K_1)}}_1\}  = \textsc{SampleProposal}(\x, l_{1:t}) $
    \For{$k = 1,\ldots,K_1$}
    \State Load $\x^{i^{(k)}}$
    \State $\mathbf{y}^{k}_{1:t} \gets \text{Glimpse}(\x^{i^{(k)}}, l_{1:t}) $
    \State Compute exact likelihood $ p(\mathbf{y}_{1:t} | \x^{i^{(k)}}, l_{1:t}) = \mathcal{N}\left( \mathbf{y}_{1:t} |
      \mathbf{y}_{1:t}^{k}, \sigma_p^2 \right) $
    \State Compute weight $ w^{(k)}_2 \gets \frac{p(\mathbf{y}_{1:t} |
      \x^{i^{(k)}}, l_{1:t})}{w^{i^{(k)}}_1} $
    \EndFor
    \State $j^{(1)}, \ldots, j^{(K_2)} \gets \textsc{Resample}(w^1_2, \ldots, w^{K_1}_2) $
    \State \textbf{return} $\x^{j^{(k)}} \text{ for } k=1,\ldots,K_2$
    \EndProcedure
  \end{algorithmic}
\end{algorithm*}

Our stochastic image retrieval procedure proceeds as follows. We begin with a
large dataset of images $\x^1,\ldots,\x^\ndata{}$ independently sampled from $p(\x)$
(or an approximation of it). We are then given some observations, $\y_{1:t},
l_{1:t}$ and, roughly speaking, want to approximate $K_2$ samples from
$p(\x|\y_{1:t}, l_{1:t})$ using $K_2$ images from the dataset. We begin by
assigning each image $i$ in the dataset a weight, $w_1^i$, which approximates
the likelihood $p(\y_{1:t}|\x^i,l_{1:t})$. This approximation is efficiently
computed using PCA, as we describe later. We then construct a categorical
proposal distribution over these images, where the probability of each is
proportional to the weight. We draw $K_1$ samples from this proposal. These
images are then retrieved from the database. To make up for the inexact weights
computed previously, we reweight these samples using exact likelihoods. These
weights are used to compute a categorical distribution over the $K_1$ images.
This distribution approximates $p(\x|\y_{1:t},l_{1:t})$ increasingly well
as $\ndata{}$ (the dataset size) and $K_1$ tend to infinity.
$K_2$ samples from this distribution are returned.

Principal component analysis allows for a memory-efficient representation of the
dataset through a mean image, $\mathbf{\mu}$, a low-dimensional vector for each
image, $\mathbf{z}^i$, and an orthogonal matrix, $\mathbf{W}$, which transforms
from an image, $\x^i$, into the corresponding $\mathbf{z}^i$ as follows:
\begin{equation}
  \mathbf{z}^i = \mathbf{W} \x^i
\end{equation}
Denoting the dimensionality of $\mathbf{z}$ as $L$, the number of images as $\ndata{}$,
and the number of pixels per image as $P$, these objects can be stored with
memory complexity $\Theta(\ndata{}L+PL)$, compared to $\Theta(\ndata{}P)$ for the entire
dataset. Since $\mathbf{W}$ is orthogonal, image $i$ can be approximated using
$\mathbf{z}^i$ as
\begin{equation}
  \mathbf{\hat{x}}^i = \mathbf{W}^\top \mathbf{z}^i
\end{equation}
and $\mathbf{\hat{x}} \approx \x$ for large enough $L$. If only
certain pixels of the reconstructed image are required, $\x^i_{p_1:p_C}$
these can be obtained efficiently by using $\mathbf{\tilde{W}}^\top$, a matrix
made up of rows $p_1$ to $p_C$ of $\mathbf{W}^\top$:
\begin{equation}
  \x^i_{p_1:p_C} = \mathbf{\tilde{W}}^\top \mathbf{z}^i.
\end{equation}
This allows us to construct an approximation of the observed portion with each
image in the dataset in a time and memory-efficient manner. Additionally, we
found that our proposal was improved when the approximate likelihood was
calculated with a reconstruction of the observations as $\mathbf{\hat{y}}_{1:t}
= \mathbf{\tilde{W}}^\top\mathbf{W} \x$, rather than the true
observations $\mathbf{y}_{1:t}$. Although this requires access to the full true
image, this is acceptable as the full image was always available when we carried
out our experimental design. Also, since this is only used to calculate the
proposal distribution, it should have limited effect on the samples returned
after new weights are calculated with the exact likelihood.

The algorithms we use in our experiments vary in the following ways from
\lref{alg:image-sampling}. They both expand the effective size of the
generated dataset by comparing the observations with a horizontally-flipped
version of each dataset image, as well as the original version. Additionally for
CelebA-HQ, in order to increase the sample size and impose an invariance to very
small translations, the likelihoods (for both the proposal and the exact
likelihood) for each image are given by summing the Gaussian likelihoods over a
grid of image patches taken at locations close to the observed location. For
CUB, a simpler alteration is made to increase the effective sample size:
$\sigma_q$ is tuned while creating each proposal distribution so that it will
have an effective sample size roughly equal to the number of samples drawn. The
attached code contains both variations.

\section{Experimental details}
\subsection{AVP-CNN training}
Here, we expand on the training procedure for the AVP-CNN which was described in
\lref{sec:nogs}. For CelebA-HQ, it was trained to predict each of the 40
attributes simultaneously by outputting a vector parameterising an independent
Bernoulli distribution for each, which we found to improve accuracy compared to
training a network for each. Similarly for CUB, the AVP-CNN output a vector of
probabilities for each of the 312 binary attributes in the dataset, in addition
to the 200-dimensional categorical distribution. These outputs were produced by
linear mappings (and a sigmoid/softmax) from the same final hidden layer. The
predictions for the binary attributes of CUB were not used; this was done purely
to improve the training of the classifier.

For both datasets, the AVP-CNN was initialised from weights pretrained on
ImageNet, with only the final layer replaced and an additional input channel
added (with weights initialised to zero). It was then trained using Adam
optimizer with a learning rate of $1\times10^{-4}$ and a batch size of 64.
Validation was performed and a checkpoint saved every epoch, and the checkpoint
which gave the lowest validation cross-entropy was used. This occurred after
$183$ epochs for the network used to create near-optimal glimpse sequences on
CUB and after $458$ for CelebA-HQ. For both datasets, the training set for the
AVP-CNN was chosen to exclude the examples on which Bayesian experimental design
was later performed in order to prevent any potential issues with using the
AVP-CNN to approximate classification distributions on data which it may be
overfitted to.

\subsection{CUB pretraining}
As mentioned in the paper, we found a pretraining stage to be important for
achieving high accuracy on CUB. During this stage, the parameters of the RNN,
glimpse embedder, and classifier were trained, while the location network was
not used. After 250 epochs of pretraining, saved parameters from the epoch with
highest validation accuracy were used for the next stage of training. These
parameters were then frozen while the location network was trained. The only
difference between our pretraining loss and our training loss is that
pretraining glimpse locations are sampled from some fixed distribution, instead
of from the distribution proposed by the location network. As mentioned in the
paper, this is a uniform distribution over all $l_t \in L$ for \RAM{}. We now
describe the heuristic distribution used to pretrain the other networks. In all
our tests, we found that this heuristic gave better performance than the uniform
distribution.

The heuristic distribution uses $\EPE_{\emptyset,\emptyset}(l)$ as a measure of
the saliency of a location. That is, the expected entropy in the posterior after
taking a single glimpse at $l$. Since this is not conditioned on any previous
glimpses, it is independent of the image being processed. This is used to create
a distribution over locations as:
\begin{equation}\label{eq:heuristic-dist}
  \log p^\text{loc}(l) = C - \gamma^{-1} \cdot \EPE_{\emptyset,\emptyset}(l)
\end{equation}
where $\gamma^{-1}$ is the inverse temperature, which we set to $1$. $C = -\log
\sum_l \exp \left( - \gamma^{-1} \cdot \EPE_{\emptyset,\emptyset}(l) \right)$ is a
constant chosen such that $p^\text{loc}$ is a normalised distribution. We
estimate $\EPE_{\emptyset,\emptyset}(l)$ for each $l$ using our BOED pipeline.

\subsection{Additional baselines} \label{ap:additional-baselines}
In addition to the hand-crafted glimpse sequences described in
\lref{sec:experiments}, we here consider another form of heuristic supervision
sequence as a baseline. These were created by sampling glimpse locations
independently at each time step from the distribution defined in
\cref{eq:heuristic-dist}. We ran experiments with both $\gamma^{-1} = 1$
(denoted \PSHO{}) and $\gamma^{-1} = 5$ (denoted \PSHF{}).

\Cref{fig:more-birds-training} shows validation accuracy for these over the
course of training. We find that networks trained with \PSHO{} and \PSHF{}
converge quickly: after 240 and 400 iterations respectively, as measured by when
they reach within 1\% of the highest validation accuracy. This may be due to the
simplicity of the policies that the supervision sequences encourage them to
learn, with identical and independent distributions over the glimpse location at
every time step. Despite their fast convergence, the validation accuracy for
these heuristics appears to be almost always lower than that for \PSNOGS{}
throughout training, and never higher by a statistically significant margin.

\subsection{Architectural details}
As mentioned, we use a learned baseline to reduce the variance of the REINFORCE
gradient estimate. Following \citet{mnih2014recurrent}, this is in the form of a
linear `baseline' network which maps from the RNN hidden state to a scalar
estimate of the reward.

The glimpse embedder for CelebA-HQ consisted of the following, in order: a
$3\times3$ convolution to 16 channels with stride 1; a ReLU activation; a
$3\times3$ convolution to 32 channels with stride 2; a ReLU and $2\times2$ max
pool; and a linear layer mapping the output of this to the $64$-dimensional RNN
input.

\subsection{Hyperparameters}
For the Monte Carlo estimation of the expected posterior entropy in
\cref{eq:approx-epe}, we use $N = 200$ Monte Carlo samples for CelebA-HQ and $N
= 100$ for CUB.

For the stochastic image completion algorithm, we use the following
hyperparameters for CelebA-HQ: $K_1 = 1000$; $K_2 = 200$; and $\sigma_p
=\sigma_q = 5 \times 2^t$ for each timestep $t$. For CUB, we use: $K_1 = 500$;
$K_2 = 100$; $\sigma_p=80$; and a dynamically adjusted $\sigma_q$ (as descibed
in \cref{ap:image-retrieval}). For both
datasets, we use a 256-dimensional latent space for the PCA and $\ndata =
1\,500\,000$.

To allow direct comparison between \PSNOGS{} and the baseline supervision
sequences (\HGS{}, \PSHO{} and \PSHF{}), we supervise the same number of
training images for each; that is, 600 for tasks on CelebA-HQ and 1000 for CUB
classification.

\section{Additional plots}
\subsection{Glimpse locations}
\begin{figure*}
  \centering
  \includegraphics[scale=1]{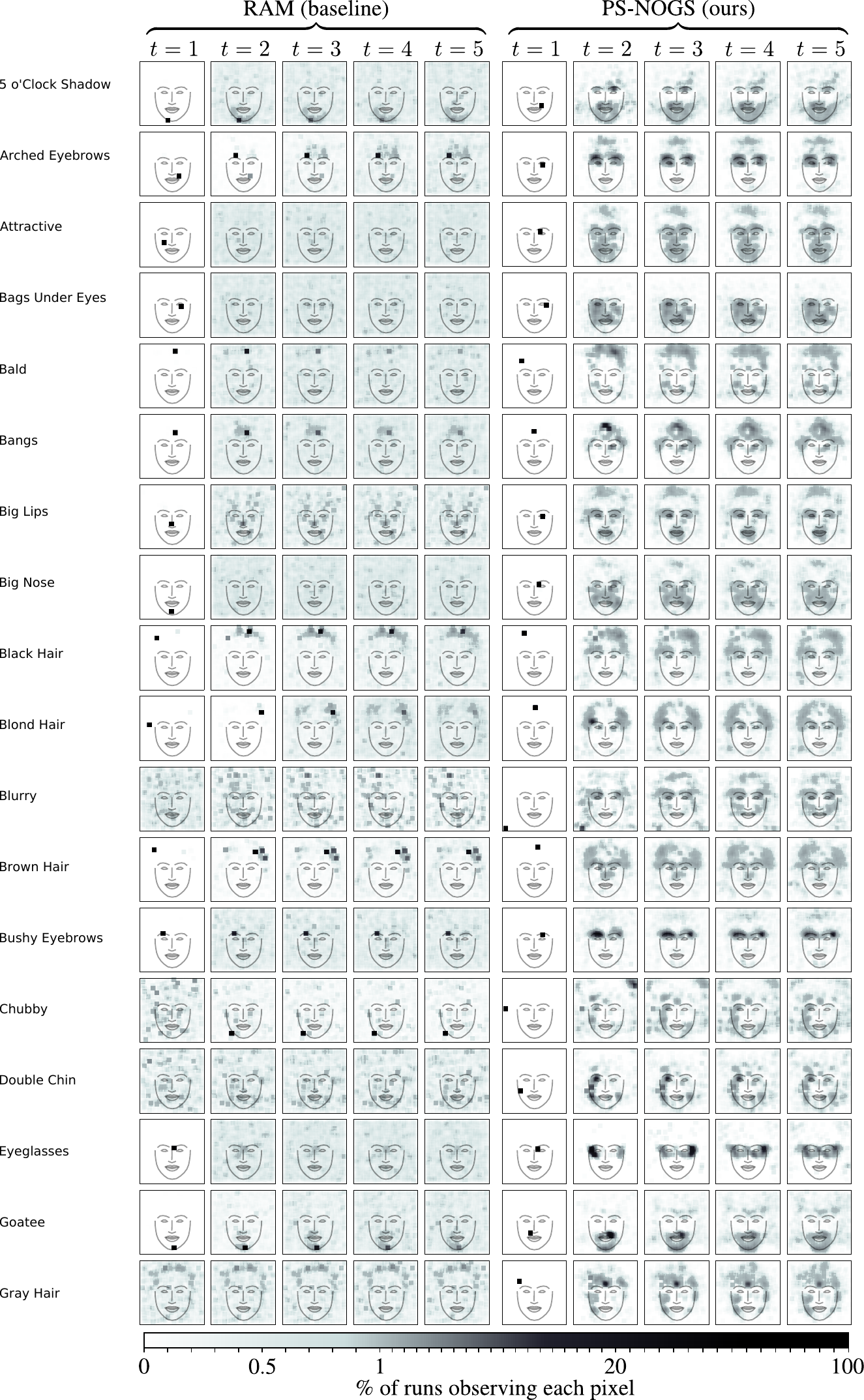}
  \caption{Glimpse locations on CelebA-HQ. }
  \label{fig:glimpses-1}
\end{figure*}

\begin{figure*}
  \centering
  \includegraphics[scale=1]{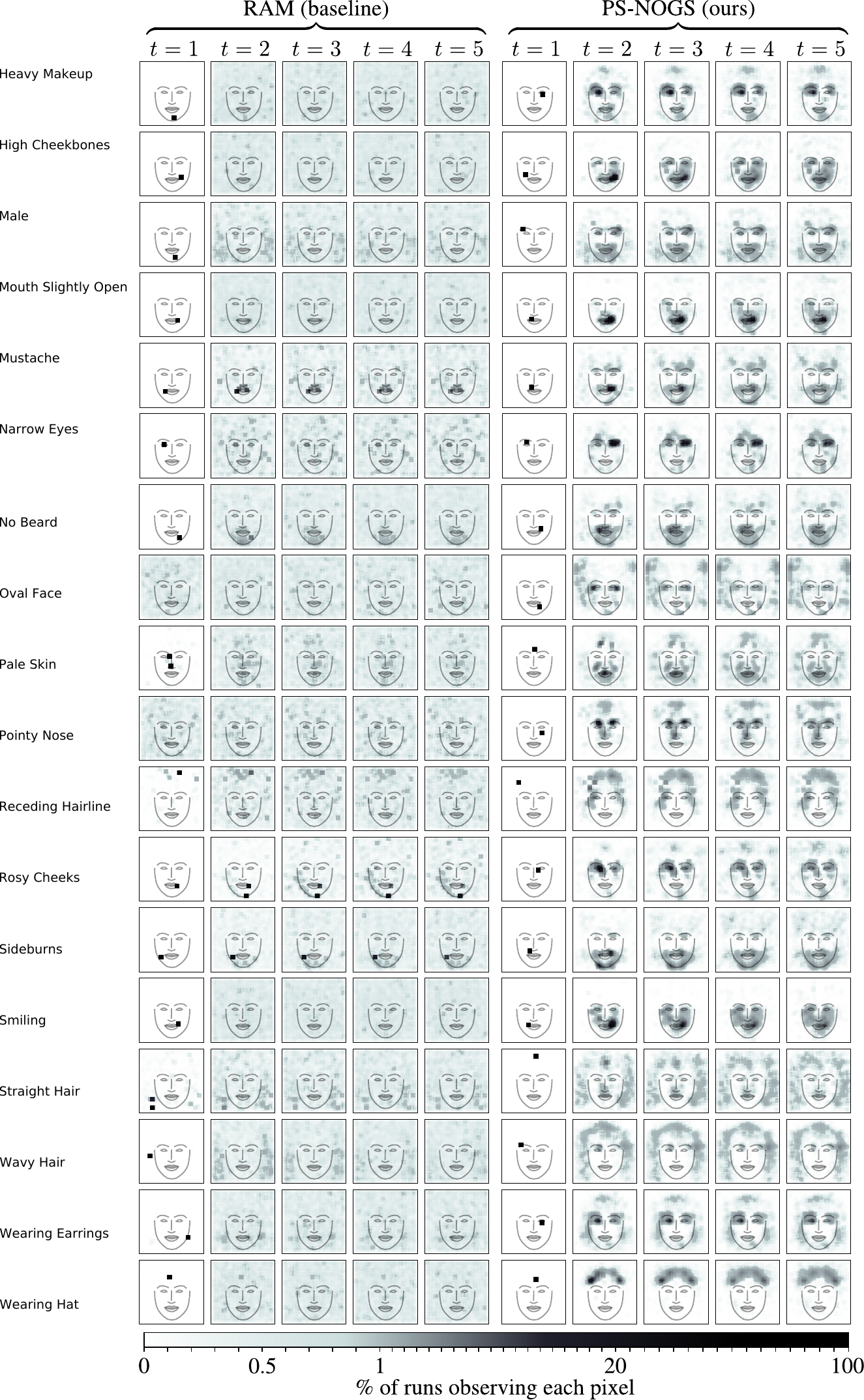}
  \caption{Glimpse locations on CelebA-HQ. }
  \label{fig:glimpses-2}
\end{figure*}

\begin{figure*}
  \centering
  \includegraphics[scale=1]{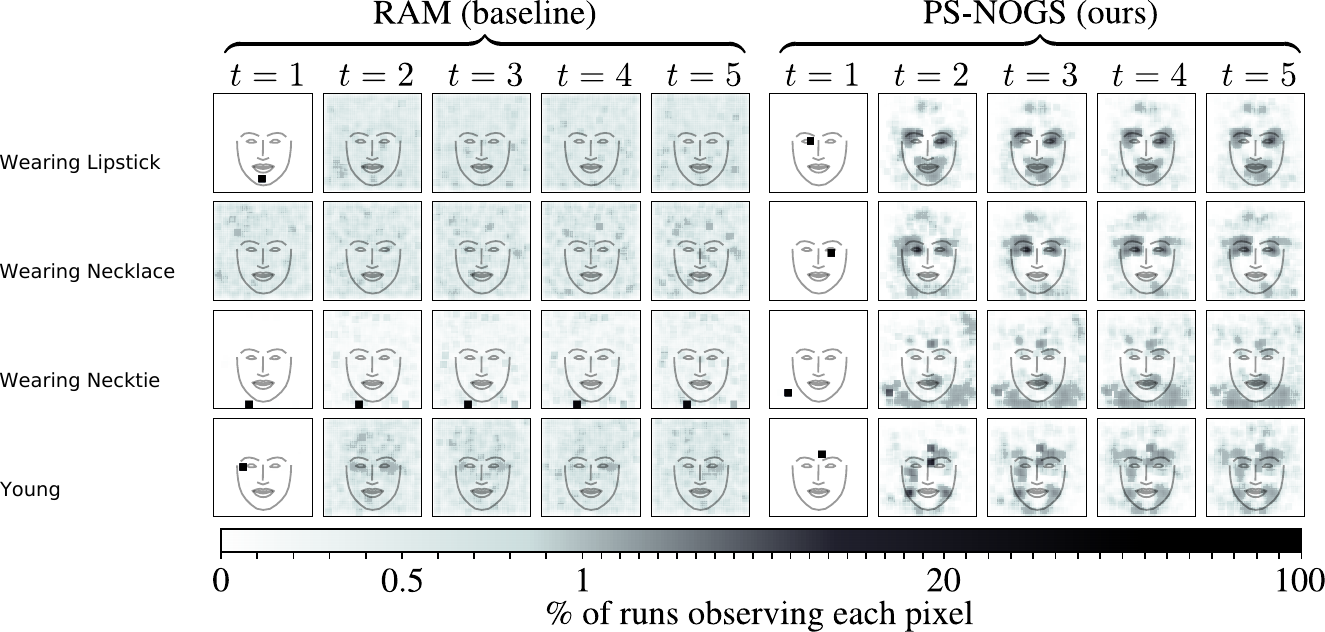}
  \caption{Glimpse locations on CelebA-HQ. }
  \label{fig:glimpses-3}
\end{figure*}

Building on \cref{fig:glimpses} in the paper, Figures \ref{fig:glimpses-1}
to \ref{fig:glimpses-3} show the glimpse locations of networks trained with
each of \RAM{} and \PSNOGS{} for all 40 CelebA-HQ attributes.
\label{ap:glimpse-policy}

\subsection{CUB training}
\begin{figure}
  \centering
  \includegraphics[scale=1]{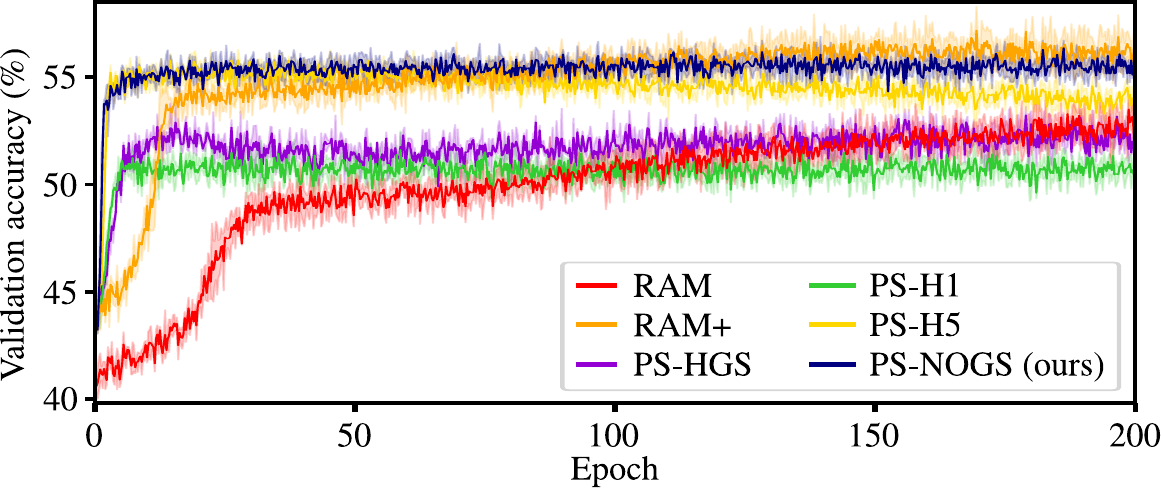}
  \caption{Expanded version of \cref{fig:birds-training} including the additional
    baselines described in \cref{ap:additional-baselines}.}
  \label{fig:more-birds-training}
\end{figure}

\Cref{fig:more-birds-training} shows the validation accuracy throughout training
for CUB, including for the additional baselines from
\cref{ap:additional-baselines}.

\subsection{Supervision sequences}
\begin{figure}
  \centering
  \includegraphics[scale=1]{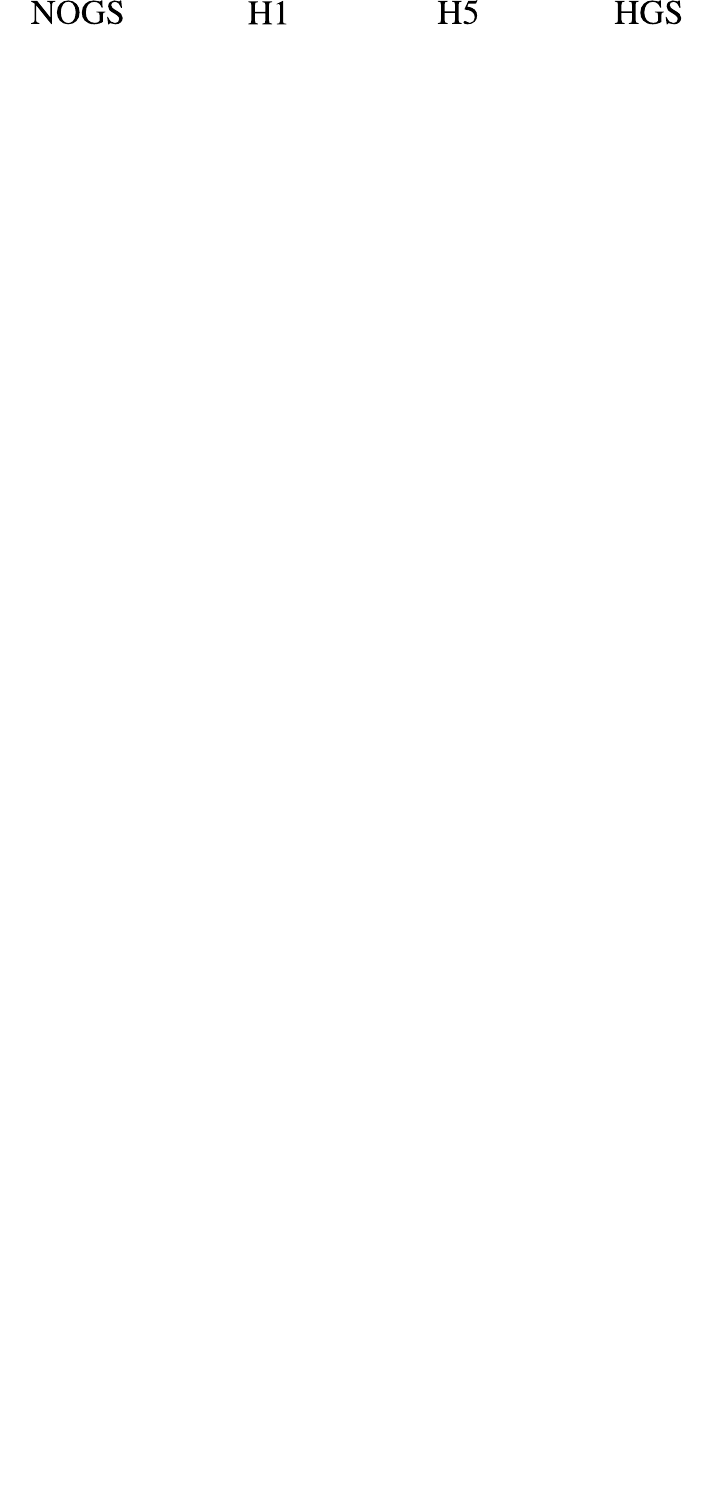}
  \caption{Examples of the image areas observed by near-optimal glimpse
    sequences (NOGS); sequences generated using the heuristic described in
    \cref{ap:additional-baselines} with $\gamma^{-1} = 1$ (H1) or $\gamma^{-1} =
    5$ (H5); and hand-crafted glimpse sequences (HGS).}
  \label{fig:supervision-seqs}
\end{figure}

\Cref{fig:supervision-seqs} shows supervision glimpse sequences given by various
methods. These are for a random sample of images.

\end{document}